\documentclass[11pt]{article}
\usepackage{acl2015}
\usepackage{times}
\usepackage{url}
\usepackage{latexsym}
\usepackage{graphicx}

\title{What Your Username Says About You}

\author{Aaron Jaech and Mari Ostendorf\\
  Dept. of Electrical Engineering \\
  University of Washington\\
  {\tt \{ajaech,ostendor\}@uw.edu} }

\date{}

\begin{document}
\maketitle
\begin{abstract}
Usernames are ubiquitous on the Internet, and they are often suggestive of user demographics.
This work looks at the degree to which gender and language can be inferred from a username alone by making
use of unsupervised morphology induction to decompose usernames
into sub-units. Experimental results on the two tasks demonstrate the 
effectiveness of the proposed morphological features compared to a character n-gram baseline.
\end{abstract}

\section{Introduction}

There is much interest in automatic recognition of demographic information of Internet users
to improve the quality of online interactions.  Researchers have looked into identifying 
a variety of factors about users, including age, gender, language, religious beliefs and political views. 
Most work leverages multiple sources of information, such as search query history, Twitter feeds, Facebook likes,
 social network links, and user profiles. However, in many situations, little of this information is available. 
Conversely, usernames are almost always available.

In this work, we look specifically at classifying gender and language based only on the username.
Prior work by sociologists has established a link
between usernames and gender \cite{cornetto}, and studies have linked
usernames to other attributes, such as individual beliefs \cite{crabill,hassa} and shown 
how usernames shape perceptions of gender and ethnicity
in the absence of common nonverbal cues \cite{pelletier}. The connections to 
ethnicity motivate the exploration of language identification.

Gender identification based on given names is very effective for English \cite{LiuRuths13}, since many names are strongly associated with a particular gender, like ``Emily'' or ``Mark''.  Unfortunately, the requirement 
that each username be unique precludes use of given names alone.
Instead, usernames are typically a combination of component words, names and numbers.
For example, the Twitter name \mbox{@taylorswift13} might decompose
into ``taylor'', ``swift'' and ``13''. The sub-units carry meaning and,
importantly, they are shared with many other individuals. Thus, our approach is to leverage 
automatic decomposition of usernames into sub-units for use in classification.

We use the Morfessor algorithm \cite{creutz2006morfessor,virpioja2013morfessor} for unsupervised 
morphology induction to learn the decomposition of the usernames into
sub-units. Morfessor has been used successfully in a variety of language modeling frameworks applied to a number of languages, particularly for learning concatenative morphological structure. The usernames that we analyze are a good match to the Morfessor framework, which allows us to push the boundary of how much can be done with only a username.  

The classifier design is described in the next section, followed by a description of experiments on gender and language recognition that  demonstrate the utility of morph-based features compared to character n-gram
features.
The paper closes with a discussion of related work and a summary of key findings.


\section{General Classification Approach}
\label{section:approach}

\subsection{Unsupervised Morphology Learning}
\label{section:morphology}

In linguistics, a morpheme is the ``minimal linguistic unit with lexical or grammatical meaning'' \cite{morphbook}.
Morphemes are combined in various ways to create longer words. Similarly, usernames
are frequently made up of a concatenated sequence of smaller units. These sub-units will be referred to as u-morphs 
to highlight the fact that they play an analogous role to morphemes but for purposes of encoding usernames rather than standard words in a language.
The u-morphs are sub-units that are small enough to be shared across different usernames but retain some meaning.

Unsupervised morphology induction using Morfessor \cite{creutz2006morfessor}
is based on a minimum description length (MDL) objective, which balances two competing goals: maximizing both the likelihood
of the data and of the model. The likelihood of the data is maximized by longer tokens and a bigger lexicon whereas 
the likelihood of the model is maximized by a smaller lexicon with shorter tokens. A parameter controls the trade-off 
between the two parts of the objective function, which 
alters the average u-morph length. We tune this parameter on held-out data
to optimize the classification performance of the demographic tasks.


Maximizing the Morfessor objective exactly is computationally intractable. The Morfessor algorithm
searches for the optimal lexicon using an iterative approach. First, the highest probability
decomposition for each training token is found given the current model. Then, the model is
updated with the counts of the u-morphs. A u-morph is added to the lexicon when it increases the weighted likelihood of the data
by more than the cost of increasing the size of the lexicon.

Usernames can be mixed-case, e.g. \mbox{``JohnDoe''}. The case change gives information 
about a likely u-morph boundary, but at the cost of doubling the size of the
character set. To more effectively leverage this cue, all characters are made lowercase but each
change from lower to uppercase is marked with a special token, e.g. ``john\$doe''. Using this encoding reduces the u-morph inventory size, and we found it to give slightly better results in language identification.

Character 3-grams and 4-grams are used as baseline features. Before extracting the n-grams
a ``\#'' token is placed at the start and end of each username. The n-grams are overlapping
to give them the best chance of finding a semantically meaningful sub-unit.

\subsection{Classifier Design}
\label{section:classification}

Given a decomposition of the username into a sequence of u-morphs (or character n-grams), we represent the relationship between the observed features and each class with a unigram language model. 
If a username $u$ has decomposition $m_1,\dots,m_n$ then it is assigned to the class $c_i$ for which the unigram
model gives it the highest posterior probability, or equivalently:
$$\mathrm{argmax}_i \ p_C(c_i)  \prod_{k=1}^n p(m_k\vert c_i),$$
where $p_C(c_i)$ is the class prior and $p(m_k\vert c_i)$ is the class-dependent
unigram.\footnote{Note that the unigram model used here, which considers only the observed u-morphs or n-grams, is not the same as using a Naive Bayes (NB) classifier based on a vector of u-morph counts. In the former, unobserved u-morphs do not impact the class-dependent probability, whereas the zero counts do impact the probability for the NB classifier. Since the vast majority of possible u-morphs are unobserved in a username, it is better to base the decision only on the observed u-morphs. The n-gram model is actually a unigram with an n-gram ``vocabulary'' rather than an n-gram language model.}


For some demographics, the class prior can be very skewed, as in the case of language detection where English is the dominant language. The choice of smoothing algorithm can be important in such cases, since minority classes have much less training data for estimating the language model and benefit from having more probability mass assigned to unseen words.  Here, we follow the approach proposed in \cite{frank2006naive} that normalizes the token count vectors for each class to have the same $L_1$ norm, specifically:
$$p(m_k\vert c_i) = \frac{1}{Z} \left( 1 + \frac{\beta \cdot n(m_k , c_i)}{n(c_i)} \right) ,$$
where $n(\cdot)$ indicates counts and $\beta$ controls the strength of the smoothing. Setting $\beta$ equal to the number of training examples approximately
matches the strength of the smoothing to the add-one-smoothing algorithm. 
 $Z=\beta + |M|$ is a constant to make the probabilities sum to one.


Only a small portion of usernames on the Internet
come with gender labels. In these situations, semi-supervised learning algorithms can use the unlabeled data 
to improve the performance of the classifier. We use a self-training expectation-maximization (EM) algorithm similar to that
described in \cite{mitchell}. 
The algorithm first learns
a classifier on the labeled data. In the E-step, the classifier assigns probabilistic labels to the unlabeled data. In the 
M-step, the labeled data and the probabilistic labels are combined to learn a new classifier. 
These steps are iterated until convergence, which usually requires three iterations for our tasks. 

Nigam et al. \shortcite{mitchell} call their method EM-$\lambda$ because it uses a parameter $\lambda$ to reduce the weight
of the unlabeled examples relative to the labeled data. This is important because the independence assumptions
of the unigram model lead to over-confident predictions. 
We used another method that directly corrects the estimated posterior probabilities. Using a small validation
set,  we binned the probability estimates and calculated the true class probability for each bin. The EM 
algorithm used the corrected probabilities for each bin for the unlabeled data during the maximization step. Samples
with a prediction confidence of less than 60\% are not used for training.



\section{Experiments}
\label{section:experiments}


\subsection{Gender Identification}

Data was collected from the OkCupid dating site by downloading up to 1,000 profiles from 27 cities in the United States, first 
for men seeking women and again for women seeking men to obtain a balanced
 set of 44,000 usernames.
The data is partitioned into three sets with 80\% assigned to training and 10\% each
 to validation and test. 
We also use 3.5M usernames from the photo messaging app Snapchat \cite{snapchat}: 1.5M are used for u-morph learning
and 2M are for self-training. 
All names in this task used only lower case, 
due to the nature of the available data.


\begin{table}[h]
\centering
\begin{tabular}{|lll|} \hline
 & {\bf Male} & {\bf Female} \\ \hline
u-morph & \begin{tabular}[c]{@{}l@{}}guy, mike, \\ matt, josh\end{tabular} & \begin{tabular}[c]{@{}l@{}}girl, marie, \\ lady, miss\end{tabular} \\ \hline
trigram & \begin{tabular}[c]{@{}l@{}}guy, uy\#,\\ kev, joe\end{tabular} & \begin{tabular}[c]{@{}l@{}}irl, gir,\\ grl, emm\end{tabular} \\ \hline
\end{tabular}
\caption{Top Gender Identification Features}
\label{table:morphfeats}
\end{table}

The top features ranked by likelihood ratios are given in \mbox{Table \ref{table:morphfeats}}. 
The u-morphs clearly carry semantic meaning, and the trigram features appear to
be substrings of the top u-morph features. The trigram features have an advantage when the u-morphs are under-segmented such
as if the u-morph ``niceguy'' or ``thatguy'' is included in the lexicon. Conversely, the n-grams can 
suffer from over-segmentation. For example, the trigram ``guy'' is inside the surname ``Nguyen''
even though it is better to ignore that substring in this context.
Many other tokens suffer from this problem, e.g. ``miss'' is in ``mission''.

The variable-length u-morphs are longer on average than the character n-grams (4.9 characters). The u-morph inventory size is similar to that for 3-grams but 5-10 times smaller than the 4-gram inventory, depending on the amount of data used since the inventory is expanded in semi-supervised training.  By using the MDL criterion in unsupervised morphology learning, the u-morphs provide a more efficient representation of usernames than n-grams and make it easier to control the tradeoff between vocabulary size and average segment length. The smaller inventory is less sensitive to sparse data in language model training.





\begin{table}[h]
\centering
\begin{tabular}{|c|rr|} \hline
 & \multicolumn{2}{c|}{Error Rate} \\ \cline{2-3}
Features & Supervised & Self-Training \\ \hline
3-gram & 28.7\% & 32.0\% \\
4-gram & 28.7\% & 29.4\% \\
u-morph & 27.8\% & 25.8\% \\ \hline
\end{tabular}
\caption{Gender Classification Results}
\label{table:genderresults}
\end{table}


The experiment results are presented in \mbox{Table \ref{table:genderresults}}. For the supervised  learning method, the
character 3-gram and 4-gram features give equivalent performance, and the u-morph features give the lowest error rate by a small amount (3\% relative).
More significantly, the character n-gram systems do not benefit from semi-supervised learning, but the u-morph features do.  The semi-supervised u-morph features obtain an error rate of 25.8\%, which represents a 10\% relative reduction over the baseline character n-gram results.



\subsection{Language Identification on Twitter}

This experiment takes usernames from the Twitter streaming API. 
Each username is associated with a tweet, for which the Twitter API identifies a language. The language labels
are noisy, so we remove approximately 35\% of the tweets where the Twitter API
does not agree with the langid.py classifier  \cite{langid}. Both training and test sets are restricted
 to the nine languages that comprise at least 1\% of the training set. 
These languages cover 96\% of the observed tweets (see Table \ref{table:langresults}).
About 110,000 usernames
were reserved for testing and 430,000 were used for training both u-morphs and the classifier. Semi-supervised methods are not used
because of the abundant labeled data. For each language, we train a one-vs.-all classifier.  The mixed case encoding technique (see  sec.~\ref{section:morphology}) gives a small increase (0.5\%) in the accuracy of the model and reduces the u-morph model size
by 5\%. 

The results in Tables \ref{table:langresults_summary} and \ref{table:langresults}  contrast systems using 4-grams, u-morphs, and a combination model, showing precision-recall trade-offs for all users together and F1 scores broken down by specific languages, respectively. The combination system simply uses the average of the posterior log-probabilities for each class giving equal weight to each model. While the overall F1 scores are similar for the 4-gram and u-morph systems, their precision and recall trade-offs are quite different, making them effective in combination. The 4-gram system has higher recall, and the u-morph system has higher precision. With the combination, we obtain a substantial gain in precision over the 4-gram system with a modest loss in recall, resulting in a 3\% absolute improvement in average F1 score. 

Looking at performance on the different languages, we find that the F1 score for the combination model is higher than the 4-gram for every language, with precision always improving. 
For the dominant languages, the difference in recall is negligible. The infrequent languages have a 4-8\% drop in recall, but the gains in precision are substantial for these languages, ranging from 50-100\% relative.
The greatest contrast between the 4-gram and the combination system can be seen for the least frequent languages, i.e.\ the languages with the least amount of training data. 
In particular, for French, the precision of the combination system (0.36) is double that of the 4-gram model (0.18) with
only a 34\% loss in recall (0.24 to 0.16).

\begin{table}[]
\centering
\begin{tabular}{|rrrr|} \hline
\multicolumn{1}{|c}{\bf Model} & \multicolumn{1}{c}{{\bf Prec.}} & \multicolumn{1}{c}{{\bf Recall}} & \multicolumn{1}{c|}{{\bf F1}} \\ \hline
4-gram & .67 & .75 & .70 \\
u-morph & .77 & .67 & .71 \\
Combination & .73 & .73 & .73 \\ \hline
\end{tabular}
\caption{Precision, recall and F1 score for language identification using the 4-gram, u-morph representations and a combination system, averaging over all users.}
\label{table:langresults_summary}
\end{table}

\begin{table}[h]
\centering
\begin{tabular}{|rrccc|} \hline
\textbf{Language} & \textbf{Freq} & \textbf{4-gr} & \textbf{u-m} & \textbf{Comb} \\ \hline
English    & 43.5 & .78  & .78 & .79 \\
Japanese   & 21.6 & .75  & .76 & .77 \\
Spanish    & 15.1 & .66  & .65 & .68 \\
Arabic     & 7.9  & .66  & .65 & .68 \\
Portuguese & 6.2  & .50  & .55 & .58 \\
Russian    & 1.8  & .40  & .58 & .45 \\
Turkish    & 1.8  & .59  & .36 & .65 \\
French     & 1.1  & .18  & .13 & .22 \\
Indonesian & 1.1  & .34  & .30 & .43 \\ \hline
\end{tabular}
\caption{Language identification performance (F1 Scores) and relative frequency in the corpus for 4-gram (4-gr) and u-morph (u-m) representations and the combination system (Comb).}
\label{table:langresults}
\end{table}


Looking at the most important features from the classifier highlights the ability of the morphemes to capture relevant meaning.
The presence of the morpheme ``juan'', ``jose'' or ``flor'' increase the probability of a Spanish language tweet by five times. The
same is true for Portuguese and the morpheme ``bieber''. The morpheme ``q8'' increases the odds of an Arabic language tweet
by thirteen times due to its phonetic similarity to the name of the Arabic speaking country Kuwait. 
Other features may simply reflect cultural norms.
For example, having an underscore in the 
username makes it five percent less likely to observe an English tweet. These highly discriminative morphemes are both long and 
short. It is hard for the fixed-length n-grams to capture this information as well as the morphemes do.

\section{Related Work}
\label{section:related}

Of the many studies on automatic classification of online user demographics, few have leveraged names or usernames at all, and the few that do mainly explore their use in combination with other features. The work presented here differs in its use of usernames alone, but more importantly in the introduction of morphological analysis to handle a large number of usernames. 

Two studies on gender recognition are particularly relevant.
Burger {\it et al.} \shortcite{burger2011discriminating} use the Twitter username (or screen name) in combination with other profile and text features to predict gender, but they also look at the use of username features alone.  The results are not directly comparable to ours, because of differences in the data set used (150k Twitter users) and the classifier framework (Winnow), but the character n-gram performance is similar to ours (21-22\% different from the majority baseline).  The study uses over 400k character n-grams (n=1-5) for screen names alone; our study indicatess that the u-morphs can reduce this number by a factor of 10. 
Burger {\it et al.} \shortcite{burger2011discriminating} used the same strategy with the self-identified full name of the user as entered into their profile, obtaining 89\% gender recognition (vs.\ 77\% for screen names). Later, Liu and Ruths \shortcite{LiuRuths13} use the full first name from a user's profile for gender detection, finding that for the names that are highly predictive of gender, performance improves by relying on this feature alone. However, more than half of the users have a name that has an unknown gender association. Manual inspection of these cases indicated that the majority includes strings formed like usernames, nicknames or other types of word concatenations.  These examples are precisely what the u-morph approach tries to address.

Language identification is an active area of research \cite{bergsma2012language,tweetLID}, but the username has not been used as a feature. Again, results are difficult to compare due to the lack of a common test set, but it is notable that the average F1 score for the combination model approaches the scores obtained on a similar Twitter language identification task where the algorithm has access to the full text of the tweet \cite{vrl2014accurate}: 73\% vs.\ 77\% .

A study that is potentially relevant to our work is automatic classification of ethnicity of Twitter users, specifically whether a user is African-American \cite{Pennacchiotti+11}. Again, a variety of content, profile and behavioral features are used. Orthographic features of the username are used (e.g.\ length, number of numeric/alpha characters), and names of users that a person retweets or replies to.  The profile name features do not appear to be useful, but examples of related usernames point to the utility of our approach for analysis of names in other fields.

\section{Conclusions}
\label{section:conclusion}

In summary, this paper has introduced the use of unsupervised morphological analysis of usernames to extract features (u-morphs) for identifying user demographics, particularly gender and language.  
The experimental results demonstrate that usernames contain useful personal information, and that the u-morphs provide a more efficient and complementary representation than character 
n-grams.\footnote{In order to allow the replicability of the experiments, software and data for building and evaluating our classifiers using pre-trained Morfessor models is available at \url{http://github.com/ajaech/username_analytics}.}
The result for 
language identification is particularly remarkable because it comes close to matching the performance achieved
by using the full text of a tweet.
The work is complementary to other demographic studies in that the username prediction can be used together with other features, both for the user and members of his/her social network. 

The methods proposed here could be extended in different directions. The unsupervised morphology learning algorithm could incorporate priors related to capitalization and non-alphabetic characters to better model these phenomena than our simple text normalization approach. More sophisticated classifiers could also be used, such as variable-length n-grams or neural-network-based n-gram language models, as opposed to the unigram model used here.  Of course the sophistication of the classifier will be limited by the amount of training data available.  

A large amount of data is not necessary to build a high precision username classifier. For example, less than 7,000 training examples
were available for Turkish in the language identification experiment and the classifier had a precision of 76\%. Since little data
is required, there may be many more applications of this type of model.

Prior work on unsupervised morphological induction focused on applying the algorithm to natural language input.
By using those techniques with a new type of input, this paper shows that there are other applications of morphology learning.


\bibliographystyle{acl}
\bibliography{morefs}

\end{document}